\documentclass[11pt]{article}
\usepackage[T1]{fontenc}
\usepackage{lmodern}

\usepackage[round]{natbib}
\usepackage{graphicx}

\usepackage[affil-it]{authblk}
\usepackage{fullpage}
\usepackage{setspace}
\newcommand{\keywords}[1]{\vspace{2mm}\noindent\textbf{\textit{Keywords:}} #1}

\usepackage{amsmath}
\usepackage{amssymb}
\usepackage{exscale}
\usepackage{mathtools}
\providecommand{\R}[0]{\mathbb{R}}
\providecommand{\relmiddle}[1]{\mathrel{}\middle#1\mathrel{}}
\providecommand{\func}[1]{\mathinner{\left({#1}\right)}}
\providecommand{\prob}[2]{\mathinner{\left({#1}\relmiddle|{#2}\right)}}
\providecommand{\expect}[2]{\mathinner{\bigl\langle #1 \bigr\rangle_{#2}}}
\providecommand{\expectsup}[3]{\mathinner{\bigl\langle #1 \bigr\rangle_{#2}^{#3}}}
\DeclareMathOperator*{\tr}{tr}

\DeclareMathOperator*{\KL}{KL}

\title{
Clustering of non-Gaussian data by variational Bayes\\
for normal inverse Gaussian mixture models
}
\author{
Takashi Takekawa\\
takekawa@cc.kogakuin.ac.jp
}
\affil{
Kogakuin University of Technology and Engineering,\\
1-24-2 Nishi-Shinjuku, Shinjuku, Tokyo 163-8677, Japan
}

\begin{document}

\maketitle

\doublespacing

\begin{abstract}
Finite mixture models, typically Gaussian mixtures,
are well known and widely used as model-based clustering.
In practical situations,
there are many non-Gaussian data
that are heavy-tailed and/or asymmetric.
Normal inverse Gaussian (NIG) distributions are normal-variance mean which mixing densities are inverse Gaussian distributions and can be used for both haavy-tail and asymmetry.
For NIG mixture models,
both expectation-maximization method
and variational Bayesian (VB) algorithms
have been proposed.
However, the existing VB algorithm for NIG mixture
have a disadvantage
that the shape of the mixing density is limited.
In this paper, we propose another VB algorithm for NIG mixture that improves on the shortcomings.
We also propose an extension of Dirichlet process mixture models to overcome the difficulty in determining the number of clusters in finite mixture models.
We evaluated the performance with artificial data and found that it outperformed Gaussian mixtures
and existing implementations for NIG mixtures,
especially for highly non-normative data.

\end{abstract}

\section{Introduction}
Finite mixture models are commonly used for density estimation or data clustering in a variety of fields
\citep{Melnykov2010, McLachlan2019}.
Finite mixture models are known
as model-based unsupervised learning
that does not use label information.
Historically, Gaussian mixture models
are most popular for model-based clustering
\citep{Celeux1995, Fraley1998}.
However,
there are many heavy-tailed and/or asymmetric cases
where normality cannot be assumed in the actual data.
Therefore, in recent years, there has been increasing attention
on the use of non-normal models in model-based clustering.
Specifically, mixture models of $t$-distributions
\citep{Shoham2002, Takekawa2009},
skew $t$-distributions \citep{Lin2007},
normal inverse Gaussian distributions \citep{Karlis2009, Subedi2014, OHagan2016a, Fang2020}
and generalized hyperbolic distributions
\citep{Browne2015} have been proposed.

For parameter estimation of the mixture distribution,
the expectation-maximization (EM) algorithm
based on the maximum likelihood inference
was classically used and is still in use today
\citep{Dempster1977}.
In the maximum likelihood method,
it is impossible to determine the number of clusters in principle.
Therefore, it is necessary to apply the EM method
under the condition of multiple number of clusters
and then determine it using some information criteria like Baysian information criteria (BIC).
Bayesian inferences make
use of prior knowledge
about clusters in the form of prior distributions.
Therefore, we can evaluate the estimation results
for different numbers of clusters based on the model evidence.
In Bayesian inference,
it is natural and common to use the Dirichlet distribution, which is a conjugate prior of the categorical distribution, as a prior for cluster concentration.
Since the Dirichlet distribution
is defined based on the number of clusters,
the disadvantage is that the prior distribution is affected by the number of clusters.
Dirichlet process mixture (DPM) models
can be used as a solution to this problem
\citep{Antoniak1974, MacEachern1994, Neal2000}.
DPM is a model that divides data
into infinite number of clusters.

There are two methods for parameter estimation
based on Bayesian inference,
one is Monte Carlo Markov chane (MCMC) sampling
and the other is variational Bayesian (VB)
\citep{Ghahramani2001, Jordan1999}.
MCMC has the advantage of being
a systematic approach to various problems.
However, it has the problem of slow convergence
and difficulty in finding convergence.
These shortcomings have a large impact
particularly on large scale problems \citep{Blei2017}.
On the other hand,
VB,
in which the independence between variables is assumed,
allow us to solve the relaxed problems faster.
VB algorithm is similar to EM algorithm,
it eliminates solves the disadvantage of the slow and unstable convergence of EM algorithm \citep{Renshaw1987}.
In addition, automatic relevance determination
eliminates unnecessary clusters
during the iteration,
and the number of clusters can be determined in a natural way \citep{Neal1996}.

Normal inverse Gaussian (NIG) distribution,
a subclass of generalized hyperbolic distributions,
is mathematically tractable and open used to treat a tail-heaviness and skewness of data.
NIG distribution is defined
as the normal variance-mean mixture
with the inverse Gaussian mixing density.
An expectation-maximization (EM) framework for mixtures of NIG was proposed by \cite{Karlis2009}.
And a VB framework for NIG mixtures
was also proposed by \cite{Subedi2014}.
Recently, \cite{Fang2020} introduced
Dirichlet process mixture to framework by \cite{Subedi2014}.
\cite{Fang2020} introduce Dirichlet process mixture models to Subedi's implementation.
However, as pointed out in this paper, the implementation of \cite{Subedi2014} and \cite{Fang2020}
have the drawback of fixing the shape of the mixing density, which represents the non-normality.

In this paper,
we introduce a approximate Bayes inference
for mixture models of NIG by VB
without fixing the shape of the mixing density.
In this formulation,
the conjugate prior of the shape of the mixing density
is a generalized inverse normal distribution,
and we propose to use inverse normal distributions
or gamma distributions as a prior,
both of these are a subclass of generalized inverse Gaussian.
For the concentration parameter,
we propose
both Dirichlet distribution model and DPM model.
Finally, the proposed method was evaluated with artificial data.
As a result, the
The proposed method is based on the non-normality of mixed distribution data.
Compared to VB for GMM and past VB for NIGMM implementations, the
Estimating the number of clusters and clustering comprehensively
The results were significantly better in terms of both the quality and aduative rank index (ARI) \citep{Hubert1985}.

\section{Methods}

In this section, we described another variational Bayes implementation
for finite mixture of NIG distributions,
in which the prior of mixing density's shape parameter $\lambda$
obeys generalized inverse Gaussian distribution.
First, the Dirichlet distribution version of VB for mixture of NIG is described in 2.1-2.4.
Then, we introduce the Dirichlet process mixture framework in 2.5.
We also discuss the policy for setting hyperparameters in 2.6.
The difference between \cite{Subedi2014} and the proposed model
is described in Appendix~B.
The details of the distributions shown in this section are described in Appendix~A.

\subsection{Multivariate normal inverse Gaussian distribution}
NIG distribution is defined as the normal variance-mean mixture
with the inverse Gaussian mixing density \citep{Barndorff-Nielsen1997}.
The mixing density $y' \in \R$ arises from an inverse Gaussian distribution $\mathcal{N}^{-1}$
with the mean $y_0 \in \R$ and the shape $\lambda \in \R$
and the observation $x \in \R^D$ arises from an $D$-dimensional multivariate normal distribution $\mathcal{N}_D$
with the mean $\mu + y' \beta' \in \R^D$ and the precision matrix $y' \tau' \in \R^{D \times D}$:
\begin{gather}
p\prob{x}{y'}
=
\mathcal{N}_D\prob{x}{\mu + y' \beta', y'^{-1} \tau'}
\quad \text{and} \quad
p\func{y'}
=
\mathcal{N}^{-1}\prob{y'}{y_0, \lambda},
\end{gather}
where $\mu \in \R^D$ and $\beta' \in \R^D$ is the center and the drift parameter, respectively.
Originally, it is assumed that $|\tau'|=1$ should be satisfied to eliminate redundancy.
However, the restriction $|\tau'|=1$ make the parameter inference difficult \citep{Protassov2004}.

\begin{figure}
    \centering
    \includegraphics[width=16cm]{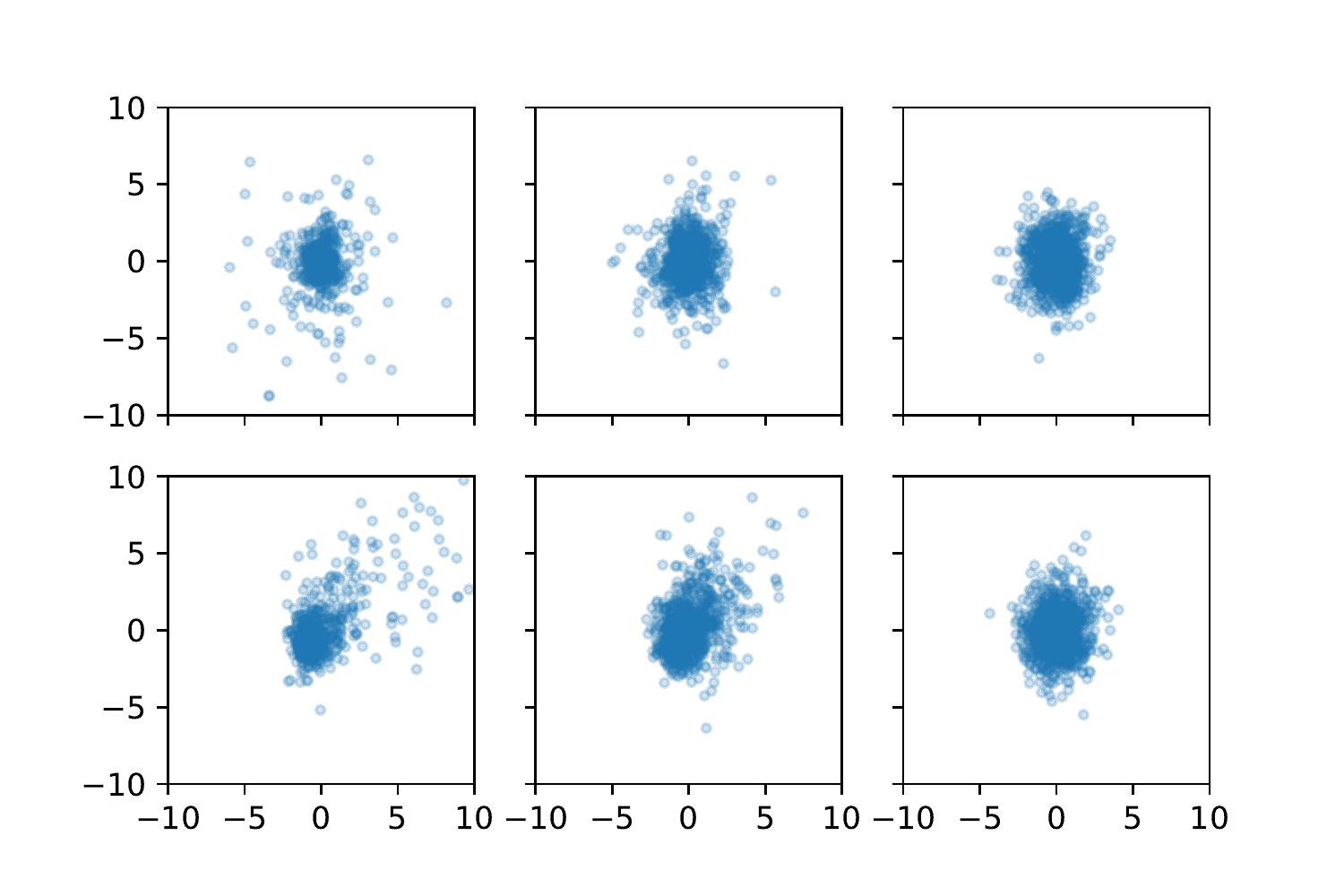}
    \caption{
    Scatter plots of samples from 2-dimensional NIG distribution:
    The precision matrix is set as $\tau=\operatorname{diag}[1, 2]^{-1}$.
    The top and bottom row correspond to bias $\beta=[0, 0]^\top$ and $\beta=[1,1]$, respectively.
    The left, center and right column correspond to normality $\lambda=0.1$, 1 and 10, respectively.
    The mean parameter is set as $\mu = -\beta$
    to satisfy the mean of the distribution is the origin.
    }
    \label{fig:nig}
\end{figure}

To avoid the difficulty,
we introduce an alternative representation which fix the mean of $\lambda$:
\begin{equation}
p\prob{x}{y}
=
\mathcal{N}\prob{x}{\mu + y \beta, y^{-1} \tau}
\quad \text{and} \quad
p\prob{y}{\lambda}
=
\mathcal{N}^{-1}\prob{y}{1, \lambda}.
\end{equation}
The representation can be easily available by the scale change $\beta = y_0 \beta'$, $\tau = y_0^{-1} \tau'$
and the property of the distribution:
\begin{equation}
\text{if} \quad
y'
\sim
\mathcal{N}^{-1}\prob{y'}{y_0, \lambda},
\quad\text{then}\quad
y = \frac{y'}{y_0}
\sim
\mathcal{N}^{-1}\prob{y}{1, \lambda}.
\end{equation}
The mean
and
the precision matrix
of normal inverse Gaussian are
$\mu + \beta$
and
$\tau$,
respectively.
The larger the normality $\lambda$,
the closer NIG distribution approaches the normal distribution.
A large bias $\beta$ results in an asymmetric distribution (see Fig.~\ref{fig:nig})

\cite{Subedi2014} also proposed a similar representation.
However, as a result, their proposal fixed the mean rather than the shape
of $\lambda$.
Then, they conclude that the conjugate prior
should obey truncated normal distributions (see Appendix B).
As a results, their representation lose the flexibility of NIG distribution.
Moreover, the redundancy and difficulty of truncated process still remain.
On the other hand, our representation could deal the entire NIG
and the conjugate prior of $\lambda$ which
obey generalized inverse Gaussian distributions
do not need additional process such as truncation.

\subsection{Variational Bayes for mixture of MNIG}

\begin{figure}
    \centering
    \includegraphics{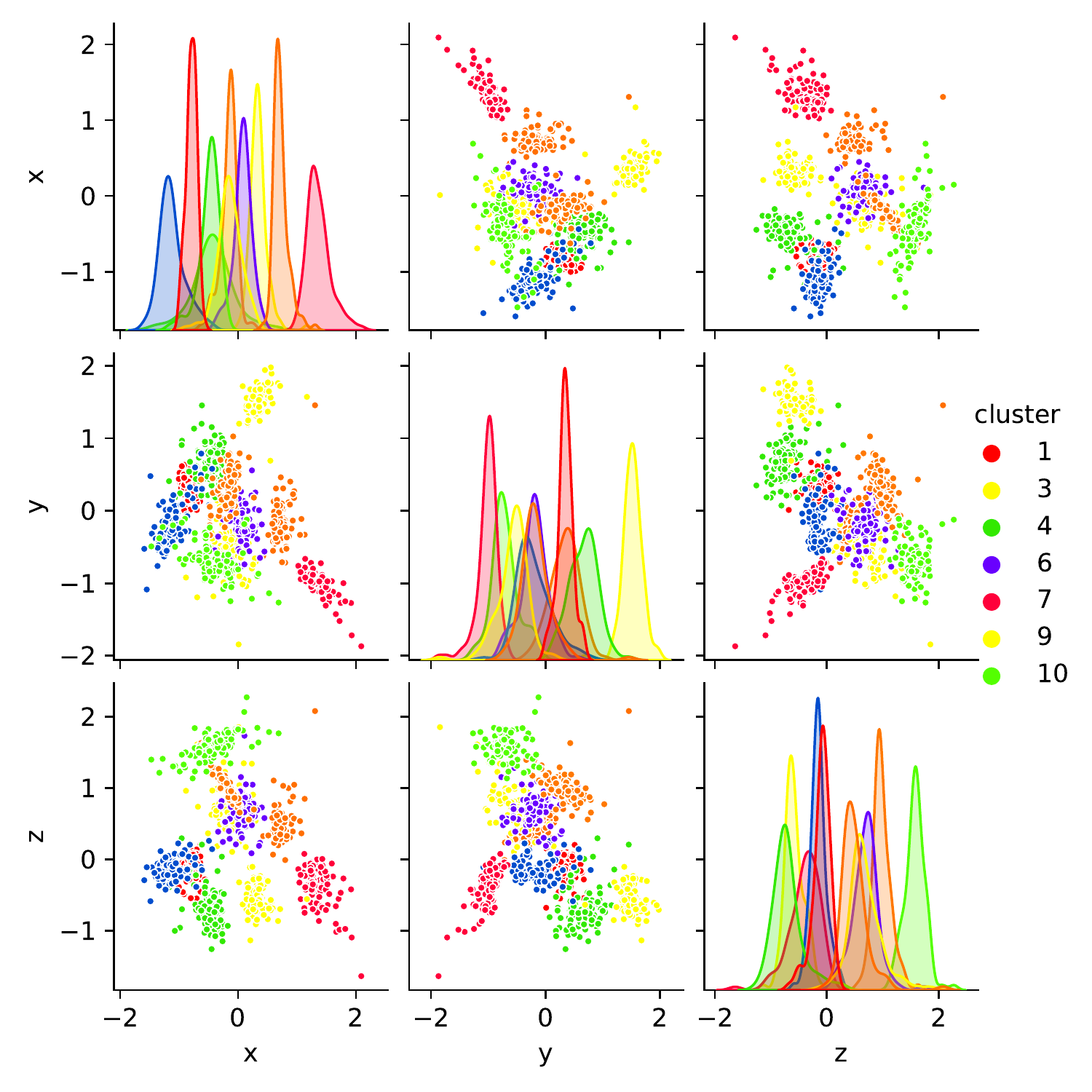}
    \caption{Example of data generated by the NIG mixture model:
    The normality parameter $\lambda$ is set around 1
    and the asymmetry parameter $\beta$ around a half of standard deviation.
    Using the parameters described in section 2.6,
    this can be describe as $\lambda_* = 1$ and $\sigma_\beta = 0.5$.
    }
    \label{fig:sample}
\end{figure}

The probability distribution function for the mixture of $M$ NIG is defined as: 
\begin{equation}
p\prob{x_i}{\alpha, \lambda, \mu, \beta, \tau}
=
\sum_{j=1}^{M}
\alpha_j
\int
\mathcal{N}^{-1}\prob{y_i}{1, \lambda_j}
\mathcal{N}\prob{x_{ij}}{\mu_j + y_i \beta_j, y_i^{-1}\tau_j}
dy_i,
\end{equation}
where $\alpha_j$ are the concentration parameters of mixture and
its satisfied
$p\left(z_i \mid \alpha\right)
=
\prod_{j=1}^{M}\alpha_j^{z_{ij}}$.
An example of the data generated by NIG mixture is shown in Fig~\ref{fig:sample}.
Here, we introduce the component indicator vector $z_i$;
$z_i = G_j$ if the subject belongs to group $j$
and $G_j$ is a one-hot encoded $D$-dimensional vector which only the $j$-th element is 1.
The joint probability of the observed data $x_i$, the mixing densities $y_i$
and the component indicators $z_i$ is described as following:
\begin{equation}
p\left(
 x_i, y_i, z_i \mid \alpha, \lambda, \mu, \beta, \tau
\right)
=
\prod_{j=1}^{M}
\left[
\alpha_j
\mathcal{N}^{-1}\left(y_i \mid 1, \lambda_j\right)
\mathcal{N}\left(x_i \mid \mu_j + y_i \beta_j, y_i^{-1}\tau_j\right)
\right]^{z_{ij}}.
\end{equation}

In the variational Bayesian (VB) algorithm,
the test function $q(y, z, \alpha, \lambda, \mu, \beta, \tau)$
which approximates the posterior $p\prob{y, z, \alpha, \lambda, \mu, \beta, \tau}{x}$ should be optimized in the sense of minimizing the KL divergence
$\KL\left[q(y, z, \alpha, \lambda, \mu, \beta, \tau), p\prob{y, z, \alpha, \lambda, \mu, \beta, \tau}{x}\right]$.
VB introduce the approximation of assuming the independence between hidden variables and parameters:
\begin{equation}
q\func{y, z, \alpha, \lambda, \mu, \beta, \tau} = q_\zeta\func{y, z} q_\theta\func{\alpha, \lambda, \mu, \beta, \tau}.
\end{equation}

In the VB algorithm, M-step update the parameter test function $q\func{\alpha, \lambda, \mu, \beta, \tau}$
by fixing the test function of hidden variable $q\func{y, z}$ the as following:
\begin{equation}\label{mstep}
    \log q\func{\alpha,\lambda,\mu,\beta,\tau}
    = \text{const} + \log p\func{\alpha,\lambda,\mu,\beta,\tau}
    +
      \expect{\log p\prob{x, y, z}{\alpha,\lambda,\mu,\beta,\tau}}{q_\zeta}.
\end{equation}
And E-step update the test function of hidden variable $q\func{y, z}$
by fixing the parameter test function $q\func{\alpha, \lambda, \mu, \beta, \tau}$ as following:
\begin{equation}\label{estep}
q\func{y_i, z_i} \propto
\exp \expect{\log p\prob{x_i, y_i, z_i}{\alpha,\lambda,\mu,\beta,\tau}}{q_\theta}.
\end{equation}
The convergence of E-step and M-step iteration can be evaluated
by the evidence lower bound (ELBO):
\begin{equation}
    L = \log p\func{x} - \KL\left[q_\theta\func{\alpha,\lambda,\mu,\beta,\tau}, p\prob{\alpha,\lambda,\mu,\beta,\tau}{x}\right].\label{elbo}
\end{equation}

\subsection{M-step}

If we have the values of the expected value of hidden variables:
\begin{gather}
    \bar{z}_{ij} = \expect{z_{ij}}{q_\zeta\left(z_i=G_j\right)},
    \quad
    \bar{y}_{ij} = \expect{y_i}{q_\zeta\prob{y_i}{z_i=G_j}},
    \quad
    \hat{y}_{ij} = \expect{y_i^{-1}}{q_\zeta\prob{y_i}{z_i=G_j}},
\end{gather}
the statics of data can be also available:
\begin{equation}
\begin{aligned}
    Z^*_j &= \sum_{i=1}^N \bar{z}_{ij} &
    Z^+_j &= \sum_{i=1}^N \bar{y}_{ij}\bar{z}_{ij}, &
    Z^-_j &= \sum_{i=1}^N \hat{y}_{ij}\bar{z}_{ij},\\
    X^*_j &= \sum_{i=1}^N \bar{z}_{ij} x_i, &
    X^-_j &= \sum_{i=1}^N \hat{y}_{ij}\bar{z}_{ij} x_i, &
    S^-_j &= \sum_{i=1}^N \hat{y}_{ij}\bar{z}_{ij} x_i x_i^\top,
\end{aligned}
\end{equation}
Using these values, the expectation term with respect to the parameters
$\alpha$, $\lambda$ , $\mu$, $\beta$ and $\tau$ in Eq.~\eqref{mstep}
can be rearranged:
\begin{multline}\label{mexpect}
\left\langle
\log p\left(x, y, z\mid\alpha,\lambda,\mu,\beta,\tau\right)
\right\rangle_{q_\zeta}
=
\mathrm{const}
+ \sum_{j=1}^M Z^*_j \log\alpha_j
\\
+ \sum_{j=1}^M \left\{
\frac{Z^*_j }{2}\log_i \lambda_j
- \frac{Z^+_j +
Z^-_j
- 2 Z^*_j}{2}
\lambda_j
- \frac{0}{2} \times \lambda_j^{-1}
\right\}
+\sum_{j=1}^M \frac{1}{2} \left\{
Z^*_j \log\det\tau_j
- \tr S^-_j \tau_j
\right\}
\\
+ \sum_{j=1}^M \frac{1}{2} \tr \left\{
- Z^-_j \mu_j \mu_j^\top
- Z^+_j \beta_j \beta_j^\top
- Z^*_j \left(\mu_j \beta_j^\top + \beta_j \mu_j^\top\right)
+ X^-_j  \mu_j^\top
+ X^*_j  \beta_j^\top
\right\}\tau_j.
\end{multline}

The prior to correspond to the form of the posterior test function could be defined as
\begin{multline}\label{prior}
\log p\left(\alpha,\lambda,\mu,\beta,\tau\right)
= \mathrm{const}
+\sum_{j=1}^M \left(l_j - 1\right) \log \alpha_j
\\
+ \sum_{j=1}^M \left\{
(h_0 - 1) \log \lambda_j - \frac{f_0}{2} \lambda_j - \frac{g_0}{2} \lambda_j^{-1}
\right\}
+ \sum_{j=1}^M \frac{1}{2} \left\{
s_0 \log\det\tau_j
- \tr t'_0 \tau_j
\right\}
\\
+ \sum_{j=1}^M 
\frac{1}{2}
\tr
\left\{
- u_0 \mu_j \mu_j^\top
- v_0 \beta_j \beta_j^\top
- w_0 \left(\mu_j \beta_j^\top + \beta_j \mu_j^\top\right)
+ m'_0 \mu_j^\top
+ n'_0 \beta_j^\top
\right\}\tau_j.
\end{multline}
Here, we can find that Eq.~\eqref{prior} represent
the combination of Dirichlet $\mathcal{D}$,
generalized inverse Gaussian $\mathcal{N}_*^{-1}$,
Wishart $\mathcal{W}$ and multivariate normal distribution as the following:
\begin{gather}
    p\func{\alpha,\lambda,\mu,\beta,\tau}
    = p\func{\alpha_1, \alpha_2, \cdots, \alpha_M} \prod_{j=1}^M p\func{\lambda_j} p\func{\tau_j} p\prob{\mu_j, \beta_j}{\tau_j}
    \label{ddp}
    ,\\
\begin{split}
    p\func{\alpha_1, \alpha_2, \cdots, \alpha_M} &= \mathcal{D}\prob{\alpha_1, \alpha_2, \cdots, \alpha_M}{l_0, \cdots, l_0},\\
    p\func{\lambda_j} &= \mathcal{N}_*^{-1}\prob{\lambda_j}{f_0, g_0, h_0}, \\
    p\func{\tau_j} &= \mathcal{W}\prob{\tau_j}{s_0, t_0}, \\
    p\prob{\mu_j, \beta_j}{\tau_j} &=
    \mathcal{N}\prob{
    \begin{matrix}
    \mu_j \\ \beta_j
    \end{matrix}
    }{
    \begin{matrix}
    m_0 \\ n_0
    \end{matrix}
    ,\;
    \begin{matrix}
    u_0\tau_j & w_0\tau_j \\ w_0\tau_j & v_0\tau_j
    \end{matrix}
    },
\end{split}\label{ps}
\end{gather}
where
\begin{equation}
\begin{gathered}
    t'_0 = t_0 + u_0 m_0 m_0^\top + w_0 m_0 n_0^\top + w_0 n_0 m_0^\top + w_0 n_0 n_0^\top,\\
    m'_0 = u_0 m_0 + w_0 n_0, \quad
    n'_0 = w_0 m_0 + v_0 n_0.
\end{gathered}
\end{equation}
The hyper parameters can be described that
$m_0$, $n_0$ and $t_0/s_0$ are the mean of $\mu$, $\beta$ and $\tau$, respectively;
$l_0$ and $s_0$ are the precision (degree of freedom) of $\alpha$ and $\tau$, respectively;
$u_0$, $v_0$ and $w_0$ are the co-variance scale of $\mu$, $\beta$ and correlation between $\mu$ and $\beta$.
We will discuss the hyper parameter for mixing density $f_0$, $g_0$, $h_0$ in subsection .

Finally, the formula to update hyper parameters of posterior is described as
\begin{gather}
    q_\theta\func{\alpha,\lambda,\mu,\beta,\tau}
    = q_\alpha\func{\alpha_1, \alpha_2, \cdots, \alpha_M} \prod_{j=1}^M q_\lambda\func{\lambda_j} q_\tau\func{\tau_j} q_\mu\prob{\mu_j, \beta_j}{\tau_j}
    \label{ddq}
    ,\\
\begin{split}
    q_\alpha\func{\alpha_1, \alpha_2, \cdots, \alpha_M} &= \mathcal{D}\prob{\alpha_1, \alpha_2, \cdots, \alpha_M}{l_1, l_2, \cdots, l_M}, \\
    q_\lambda\func{\lambda_j} &= \mathcal{N}_*^{-1}\prob{\lambda_j}{f_j, g_j, h_j}, \\
    q_\tau\func{\tau_j} &= \mathcal{W}\prob{\tau_j}{s_j, t_j}, \\
    q_\mu\prob{\mu_j, \beta_j}{\tau_j} &=
    \mathcal{N}\prob{
    \begin{matrix}
    \mu_j \\ \beta_j
    \end{matrix}
    }{
    \begin{matrix}
    m_j \\ n_j
    \end{matrix}
    ,\;
    \begin{matrix}
    u_j\tau_j & w_j\tau_j \\ w_j\tau_j & v_j\tau_j
    \end{matrix}
    }.
\end{split}\label{qs}
\end{gather}
where
\begin{equation}
\begin{gathered}
    l_j = l_0 + Z^*_j,\quad
    f_j = f_0 + \frac{1}{2} Z^*_j,\quad
    g_j = g_0 + Z^-_j + Z^+_j - 2 Z^*_j,\quad
    h_j = h_0,\\
    s_j = s_0 + Z^*_j,\quad
    t_j = t'_0 + S^-_j,\quad
    u_j = u_0 + Z^-_j,\quad
    v_j = v_0 + Z^+_j,\quad
    w_j = w_0 + Z^*_j,\\
    u_j m_j + w_j n_j = m'_0 + X^-_j,\quad
    w_j m_j + v_j n_j = n'_0 + X^*_j.
\end{gathered}
\end{equation}
The hyper parameters of the test
function
$l_j$, $f_j$, $g_j$, $h_j$, $s_j$,
$t_j$, $u_j$, $v_j$ and $w_j$
are the sum of the prior hyper parameter
and the statistical value of observed and hidden variables.
The hyper parameter of the mean $m_j$ and bias $n_j$ can be calculated as
\begin{equation}
    \begin{bmatrix}
        m_j\\
        n_j
    \end{bmatrix}
    =
    \begin{bmatrix}
        u_j & w_j \\ w_j & v_j
    \end{bmatrix}^{-1}
    \begin{bmatrix}
        m'_0 + X^-_j\\
        n'_0 + X^*_j
    \end{bmatrix}
    =
    \begin{bmatrix}
            \frac{v_j}{u_j v_j - w_j^2} & 
            \frac{-w_j}{u_j v_j - w_j^2} \\
            \frac{-w_j}{u_j v_j - w_j^2} & 
            \frac{u_j}{u_j v_j - w_j^2}
    \end{bmatrix}
    \begin{bmatrix}
        m'_0 + X^-_j\\
        n'_0 + X^*_j
    \end{bmatrix}
    .
\end{equation}

\subsection{E-step}

By calculating the expectations and organizing for $y$ in Eq.~\eqref{estep},
we obtain the following equation:
\begin{equation}\label{estep2}
\expect{\log p\prob{x_i, y_i, z_i}{\alpha,\lambda,\mu,\beta,\tau}}{q_\theta}
= \sum_{j=1}^{M} z_{ij} \left[\log \rho_{ij}
+ \log \mathcal{N}_*^{-1}\prob{y_j}{a_j, b_{ij}, c}
\right],
\end{equation}
where $c = -\frac{D+1}{2}$,
\begin{equation}
\begin{aligned}
a_j &=
\expect{\lambda_j}{q_\theta}
+ \tr \expect{\tau_j\beta_j\beta_j^\top}{q_\theta}
+ \tr \expect{\tau_j}{q_\theta}\expect{\beta_j}{q_\theta}
\expectsup{\beta_j}{q_\theta}{\top}
,\\
b_{ij} &=
\expect{\lambda_j}{q_\theta}
+ \tr \expect{\tau_j\mu_j\mu_j^\top}{q_\theta}
+ \tr \expect{\tau_j}{q_\theta}
\func{x_i-\expect{\mu_j}{q_\theta}}
\func{x_i-\expect{\mu_j}{q_\theta}}^\top
,\label{ab}
\end{aligned}
\end{equation}
and
\begin{multline}
    \log \rho_{ij} =
- \frac{D+1}{2}\log 2\pi
+ \expect{\log\alpha_j}{q_\theta}
+ \frac{1}{2}\expect{\log\lambda_j}{q_\theta}
+ \expect{\lambda_j}{q_\theta}
+ \frac{1}{2} \expect{\log\det\tau_j}{q_\theta}
\\
- \tr \expect{\tau_j\mu_j\beta_j}{q_\theta}
+ \tr \expect{\tau_j}{q_\theta}\func{x_i-\expect{\mu_j}{q_\theta}}\expectsup{\beta_j}{q_\theta}{\top}
- \log \Delta\func{a_j, b_{ij}, c}
.\label{rho}
\end{multline}
The integral constant of generalized 
inverse Gaussian distribution $\Delta\func{a_j, b_{ij}, c}$
and the expectations of parameters are described in Appendix~A.

From Eq.~\eqref{estep} and \eqref{estep2},
the test function of hidden variables can be
written with generalized inverse Gaussian and
categorical distributions:
\begin{align}
q_\zeta\func{y_i, z_i} &= q_\zeta\prob{y_i}{z_i} q_\zeta\func{z_i},
&
q_\zeta\prob{y_i}{z_i=G_j} &= \mathcal{N}_*^{-1}\prob{y_i}{a_j, b_{ij}, c},
&
q_\zeta\func{z_i=G_j} &\propto \rho_{ij}.
\end{align}
We can finally calculate expectation value of $y$, $z$ which is used in M-step:
\begin{align}
    \bar{z}_{ij}
    &= \frac{\rho_{ij}}{\sum_{j'=1}^M\rho_{ij'}}
    ,
    &
    \bar{y}_{ij}
    &= \expect{y_j}{\mathcal{N}_*^{-1}\prob{y_j}{a_j, b_{ij}, c}}
    ,
    &
    \hat{y}_{ij}
    &= \expect{y_j^{-1}}{\mathcal{N}_*^{-1}\prob{y_j}{a_j, b_{ij}, c}}
    .
\end{align}

\subsection{Dirichlet process mixtures}

In Dirichlet process mixture models,
the concentration parameters $\alpha$ can be represented by the stick-breaking process
using the collections of independent random variables $\gamma$ as follow:
\begin{equation}\label{gamma}
    \alpha_j = \gamma_j \prod_{j'=1}^{j-1} \func{1-\gamma_{j'}}.
\end{equation}
Then, the term corresponding to $\alpha$ in Eq.~\eqref{prior}
can be re-writen by $\gamma$ as 
\begin{equation}\label{dpm}
\begin{split}
\sum_{j=1}^M Z^*_j \log \alpha_j
&=
\sum_{j=1}^M Z^*_j \left[\log \gamma_j + \sum_{j'=1}^{j-1} \log \func{1-\gamma_{j'}}\right]\\
&= 
\sum_{j=1}^M \left[ Z^*_j \log \gamma_j + \sum_{j'=j+1}^M Z^*_{j'} \log \func{1-\gamma_j} \right]
\end{split}.
\end{equation}
Since Eq.~\eqref{dpm} consist of $\log \gamma_j$ and $\log \func{1-\gamma_j}$,
the conjugate prior of $\gamma_j$ should be beta distributions $\mathcal{B}$.
The prior and test function in the case of Dirichlet distribution
which described in Eq.~\eqref{ddp} and \eqref{ddq}
are replaced for DPM by
\begin{gather}
    p\func{\gamma,\lambda,\mu,\beta,\tau}
    = \prod_{j=1}^M p\func{\gamma_j} p\func{\lambda_j} p\func{\tau_j} p\prob{\mu_j, \beta_j}{\tau_j}
    \\
    \intertext{and}
    q_\theta\func{\gamma,\lambda,\mu,\beta,\tau}
    = \prod_{j=1}^M q\func{\gamma_j} q\func{\lambda_j} q\func{\tau_j} q\prob{\mu_j, \beta_j}{\tau_j},
\end{gather}
where
\begin{align}
    p\func{\gamma_j} &= \mathcal{B}\prob{\gamma_j}{l_0, r_0},
    &
    q_\gamma\func{\gamma_j} &= \mathcal{B}\prob{\gamma_j}{l_j, r_j}.\label{qgamma}
\end{align}
$p\func{\lambda_j}$, $p\func{\tau_j}$, $p\prob{\mu_j, \beta_j}{\tau_j}$,
$q\func{\lambda_j}$, $q\func{\tau_j}$ and $q\prob{\mu_j, \beta_j}{\tau_j}$
are the same in Eq.~\eqref{ps} and \eqref{qs}.

The M-step for DPM is the same as the M-step for Dirichlet distribution model
expect for the update rule of $r$:
\begin{align}
    l_j &= l_0 + Z^*_j, 
    &
    r_j &= r_0 + \sum_{j'=j+1}^M Z^*_{j'}.
\end{align}

In the E-step for DPM,
the only difference from Dirichlet distribution models
is the expectation values of $\log \alpha$ in Eq.~\eqref{rho}:
\begin{equation}\label{dpmla}
    \expect{\log \alpha_j}{q_\theta}
    =
    \expect{\log \gamma_j}
    {\mathcal{B}\prob{\alpha_j}{l_j, r_j}}
    + \sum_{j'=1}^{j-1}
    \expect{\log \func{1-\gamma_{j'}}}
    {\mathcal{B}\prob{\gamma_{j'}}{l_{j'}, r_{j'}}}.
\end{equation}

\subsection{Priors}
In this paper, we define the prior by the hyper-parameters
using the mean $\mu_x$ and co-variance matrix $\Sigma_x$ of data.
The mean of cluster centers is same as the center of data; $m_0 = \mu_x$.
The mean of bias is zero vector; $n_0 = 0$.
Basically, an uninformed prior is defined for the concentration parameter $\alpha$:
$l_0 = 1$ for Dirichlet distribution and $l_0 = r_0 = 1$ for Dirichlet process mixtures.
Increasing $l$ in the case of DD and $r$ in the case of DPM
favors small clusters and increases the overall number of clusters.

\begin{figure}
    \centering
    \includegraphics{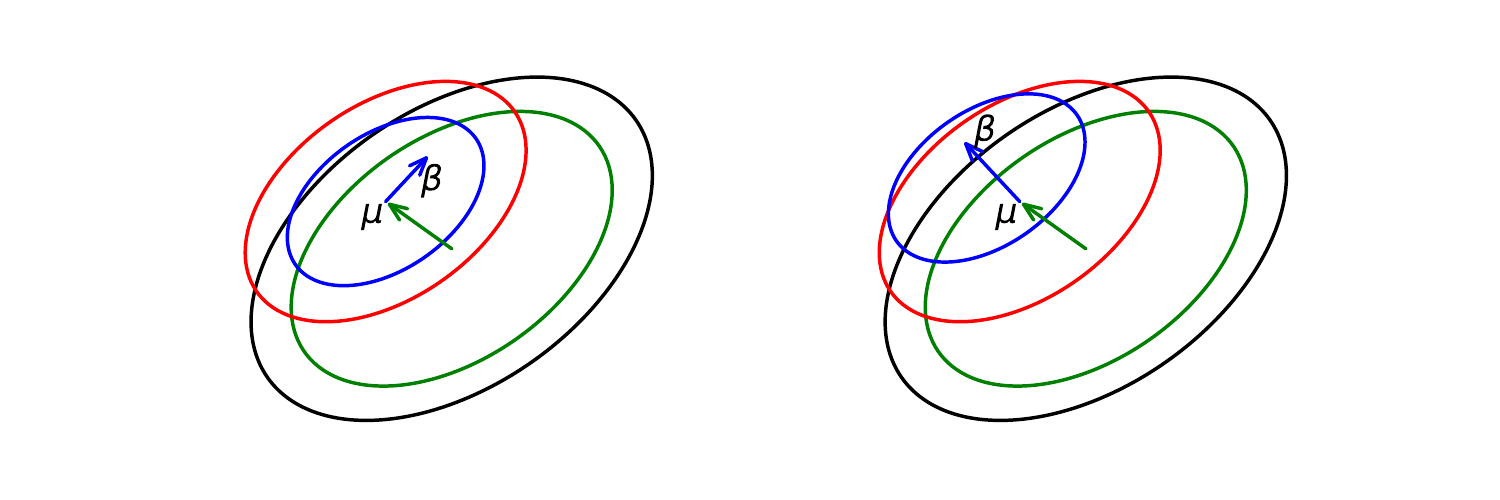}
    \caption{
Prior setting:
Centers $\mu$ has a high probability
of being inside the green ellipse
which is $\eta_\mu$ times data range (black).
Red ellipse indicate the cluster
and it is $\eta_\tau$ times data range.
Bias $\beta$ has a high probability
of being inside the blue ellipse
with center $\mu$ and $\eta_\beta$ times cluster range.
If $\zeta$ is positive, the center of $\beta$
shifts to the direction of $\mu$.
}
    \label{fig:prior}
\end{figure}

Here, we consider how to set parameters
that reflect the structure of the data as much as possible.
An overview of the structure is shown in Fig.\ref{fig:prior}.
We first assume that $\eta_\tau$ is the ratio of the size of the cluster defined
by the co-variance matrix $\tau^{-1}$ to the size of the whole data defined by $\Sigma_x$:
\begin{equation}
    \expect{\tau_j}{p}^{-1}
    = s_0^{-1} t_0 = \eta_\tau^2\Sigma_x,
\end{equation}
The range of $\mu$ present is shown as a ratio $\eta_\mu$ to the total data $\Sigma_x$,
and the range of $\beta$ is shown as a ratio $\eta_\beta$ to $\tau$.
The correlation between $\mu$ and $\beta$ is defined by $\xi \in [-1,1]$.
In other words, from the prior distribution of $\alpha$ and $\beta$,
the following equation holds:
\begin{equation}
    \begin{bmatrix}
    u_0 \expect{\tau_j}{p} & w_0 \expect{\tau_j}{p} \\
    w_0 \expect{\tau_j}{p} & v_0 \expect{\tau_j}{p}
    \end{bmatrix}^{-1}
    =
    \begin{bmatrix}
    \eta_\mu^2 \Sigma_x & \eta_\mu\eta_\beta\xi \Sigma_x^\frac{1}{2}\expectsup{\tau_j}{p}{-\frac{1}{2}}\\
    \eta_\mu\eta_\beta\xi\Sigma_x^\frac{1}{2}\expectsup{\tau_j}{p}{-\frac{1}{2}}
    & \eta_\beta^2 \expect{\tau_j}{p}
    \end{bmatrix}.
\end{equation}
Finally, we set the degree of freedom (confidence level) of $\tau$
as $s_0 = \nu_\tau$.
To summarize the above equations,
$s_0$, $t_0$, $u_0$, $w_0$ and $v_0$ can be expressed
using $\eta_\mu$, $\eta_\tau$ and $\eta_\beta$,
\begin{align}
s_0 &= \nu_\tau, &
t_0 &= \nu_\tau \eta_\tau^2 \Sigma_x, &
u_0 &= \frac{\eta_\tau^2}{\eta_\mu^2\func{1-\xi^2}}, &
w_0 &= \frac{\eta_\tau\xi}{\eta_\mu\eta_\beta\func{1-\xi^2}}, &
v_0 &= \frac{1}{\eta_\beta^2\func{1-\xi^2}}.\label{stuwv}
\end{align}

For the shape of mixing density (normality) is $\lambda_0$,
the mean $\lambda_0 = \expect{\lambda}{p}$
and the shape
$\nu_\lambda = \lambda_0^2 \expectsup{\func{\lambda-\lambda_0}^2}{p}{-1}$
are used to define the hyper-parameters.
The conjugate prior is generalized inverse Gaussian,
but its special cases inverse Gaussian and Gamma were used for the prior distribution.
The hyper-parameters of $\lambda$ are defined for inverse Gaussian prior as
\begin{gather}
f_0 = \nu_\lambda \lambda_0^{-1}
,\quad
g_0 = \nu_\lambda \lambda_0
,\quad
h_0 = -\frac{1}{2}
.\label{fgh0}
\end{gather}
On the other hand,
Gamma distribution with the mean $\lambda_0$ and shape $\nu_\lambda$
can be defined as
\begin{gather}
f_0 = 2\nu_\lambda \lambda_0^{-1}
,\quad
g_0 = 0
,\quad
h_0 = \nu_\lambda
.\label{fgh1}
\end{gather}

In this paper,
we basically set $\eta_\mu=1$, $\eta_\tau=0.3$, $\eta_\beta=0.3$, $\xi=0$,
$\lambda_0=5$, $\nu_\tau=D+1$ and $\nu_\lambda=1$.
Since the spatial size of the cluster and the nature of the bias
varies from data to data,
it is useful to set the parameters appropriately.
However, setting $nu_\tau$ to a small value
can reduce the influence of the parameters.
If the nature of the data is actually known,
a larger $nu_\tau$ will give more appropriate results.
Similarly, for normality $\lambda$,
it is important to set appropriately
$\lambda_0$ and its influence $\nu_\lambda$.

\subsection{Initial and convergence conditions}

The initial conditions are set to $\bar{y}=\hat{y}=1$,
with $z$ being the one hot representation
based on the clusters obtained by the K-means algorithm.
Then, apply M-step first, then the E-step.
If the estimated number in cluster $Z_j^*$ shrinks
less than $\varepsilon_z=2$ during the iteration,
the corresponding cluster is removed and the algorithm proceeds.

If the change in ELBO $L$ is smaller than $\varepsilon_{dL}=10^{-5} N$ five times in a row,
algorithm is terminated.
After finding $\rho_{ij}$ in E-step,
ELBO is evaluated by the following equation \citep{Takekawa2009}:
\begin{equation}
    L = \sum_i \log \sum_j \rho_{ij}
    - \KL\left[q_\theta\func{\alpha,\lambda,\mu,\beta,\tau}, p\func{\alpha,\lambda,\mu,\beta,\tau}\right].
\end{equation}

\section{Results}

\begin{figure}
    \centering
    \includegraphics{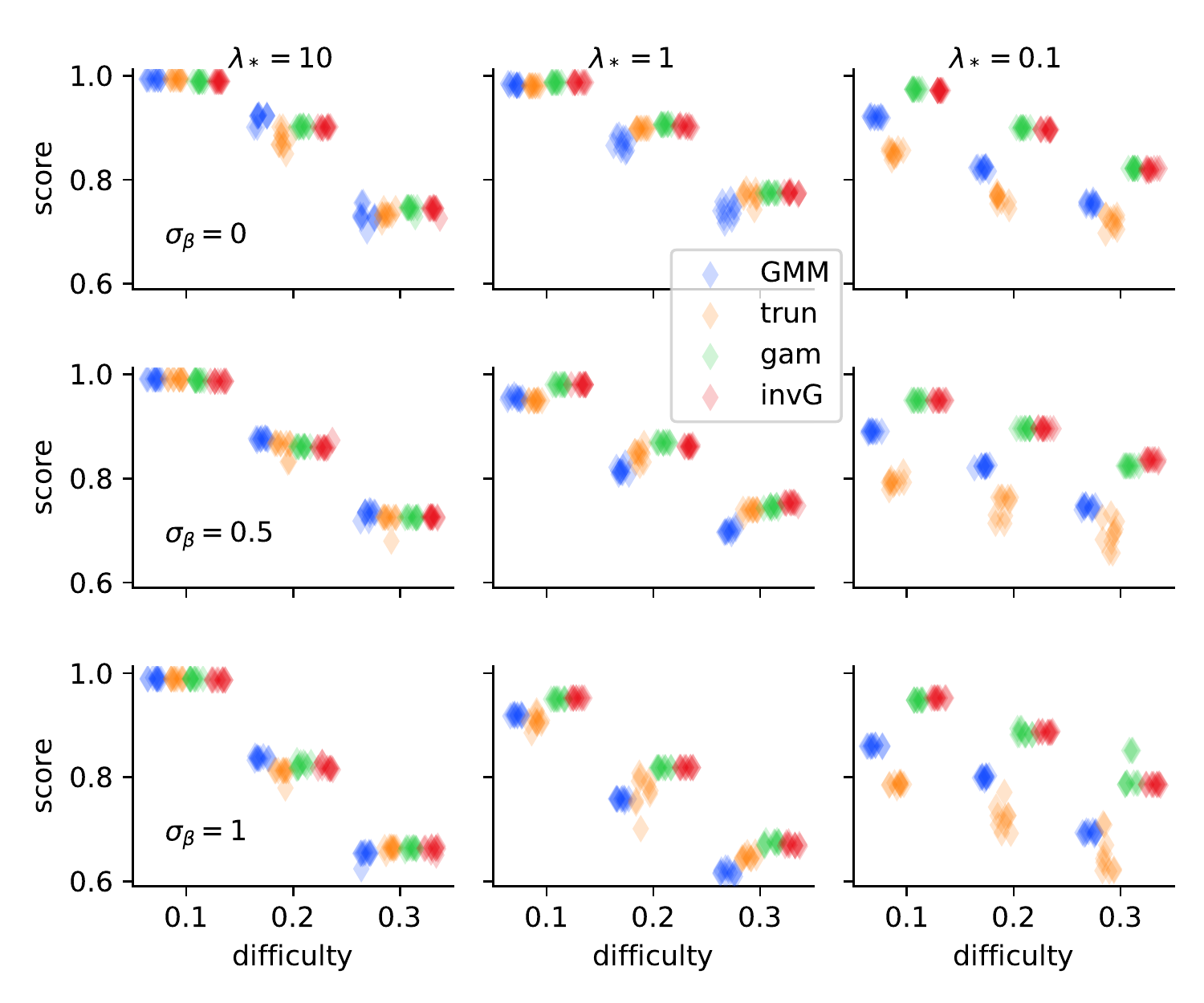}
    \caption{ARI score of simulation data.}
    \label{fig:score}
\end{figure}

\begin{figure}
    \centering
    \includegraphics{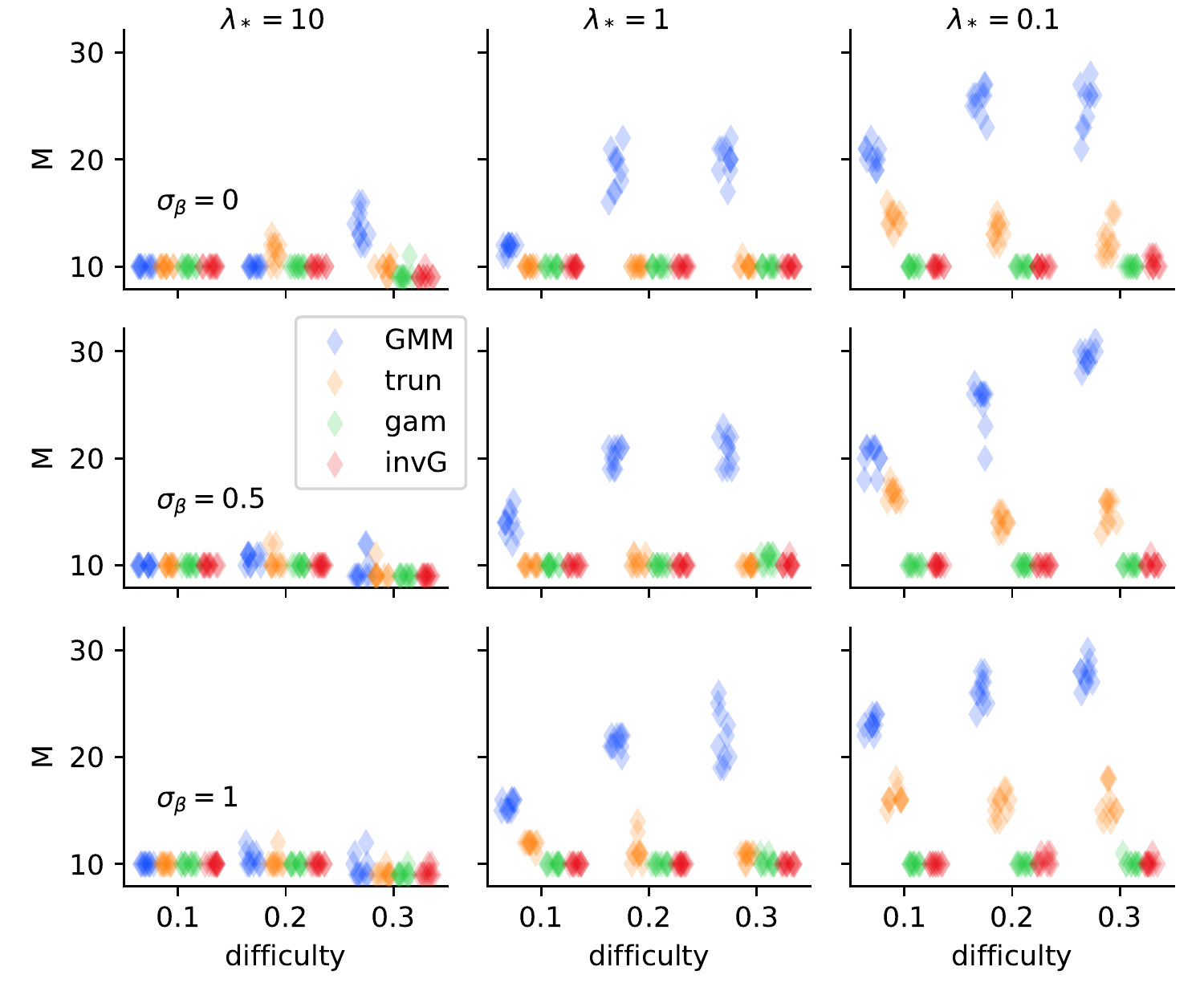}
    \caption{Estimate number of clustering. The truth is $M=10$.}
    \label{fig:num}
\end{figure}

As simulation data, the
$M$ centers $\mu_1, \cdots, \mu_M$ are generated from the normal distribution of the average 0
and covariance matrix $I$.
Similarly, we generate the bias $\beta_1, \cdots, \beta_M$
from the normal distribution of mean 0 and covariance matrix $\sigma_\beta^2I$.
The precision matrix $\tau_1, \cdots, \tau_M$ are also generated
from a Wishart distribution with mean $\sigma^{-2}I$ and degrees of freedom $D+5$.
The normalities $\lambda_1, \cdots, \lambda_M$ are generated
from an inverse normal distribution with mean $\lambda_*$ and shape parameter 5.
Finally, a sample data with $N=1000$ is generated 's using the above parameters.
The number of data in the cluster was prepared for two cases:
the uniform case and the non-uniform case.
In the uniform case, each cluster contains 100 data.
In the non-uniform case,
there are two large clusters with 400 and 200 data
and eight small clusters with 50 data.
An example of 3D data generated in Fig.~\ref{fig:sample} is shown.

In the following, we control the normality $\lambda_j$
by $\lambda_*$ and the asymmetry by $\sigma_\beta$.
In addition, we adjust the difficulty
by approaching the relative distance between clusters by $\sigma$.
Basically, algorithms are applied 10 times
par dataset with different initial conditions.
Initial number of the cluster $M_0$ is set to 50.
We evaluate the performance of the clustering using ARI.
Hereafter, we name that 
VB for Gaussian mixture models as GMM,
VB for NIG mixture models with $\lambda$ shape fixed as trun,
the proposed model with gamma prior as gam
and the proposed model with inverse Gaussian prior as invG,
respectively.

For the case of high normality ($\lambda_* = 10$; see left columns of Fig.~\ref{fig:score} and \ref{fig:num}),
the results of four algorithms were almost identical.
Although the ARI decreased with increasing difficulty (Fig.~\ref{fig:score}),
the estimate of the number of clusters was generally close to correct answer $M=10$ (Fig.~\ref{fig:score}).
For the case of $\lambda_* = 1$, the ARI of the proposed models (gam and invG)
are slightly higher than the ARI of GMM and trun.
This tendency is especially strong when the asymmetory $\sigma_\beta$ is large.
In most cases, GMM fails to estimate the number of clusters.
For the case of highly non-normal and tail-heavy ($\lambda_* = 0.1$),
the ARI of the proposed models (gam and invG)
are significantly higher than the ARI of GMM and trun.
In particular, the results of trun have a large variation and low values.
This is because trun assumes that $\lambda=1$.
It can be interpreted as not being able to cope with different situations than expected.
Overall, the proposed model estimates the correct number of clusters
in all cases and obtains a high ARI score.
The proposed method showed higher AIR and less variability especially when the normality was lower.

\begin{figure}
    \centering
    \includegraphics{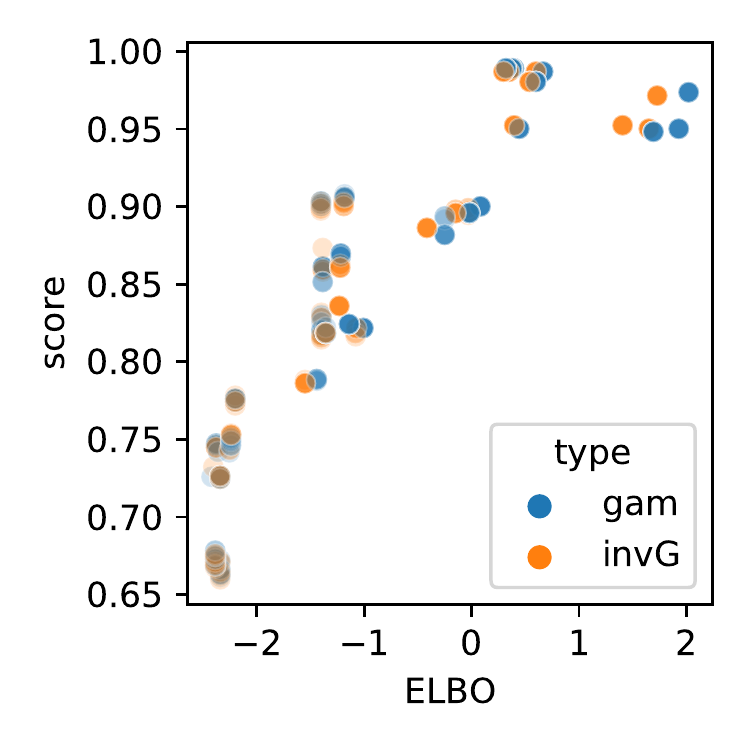}
    \caption{
    Relationship between ELBO and ARI in the proposed algorithms.
    }
    \label{fig:lb}
\end{figure}

\begin{figure}
    \centering
    \includegraphics{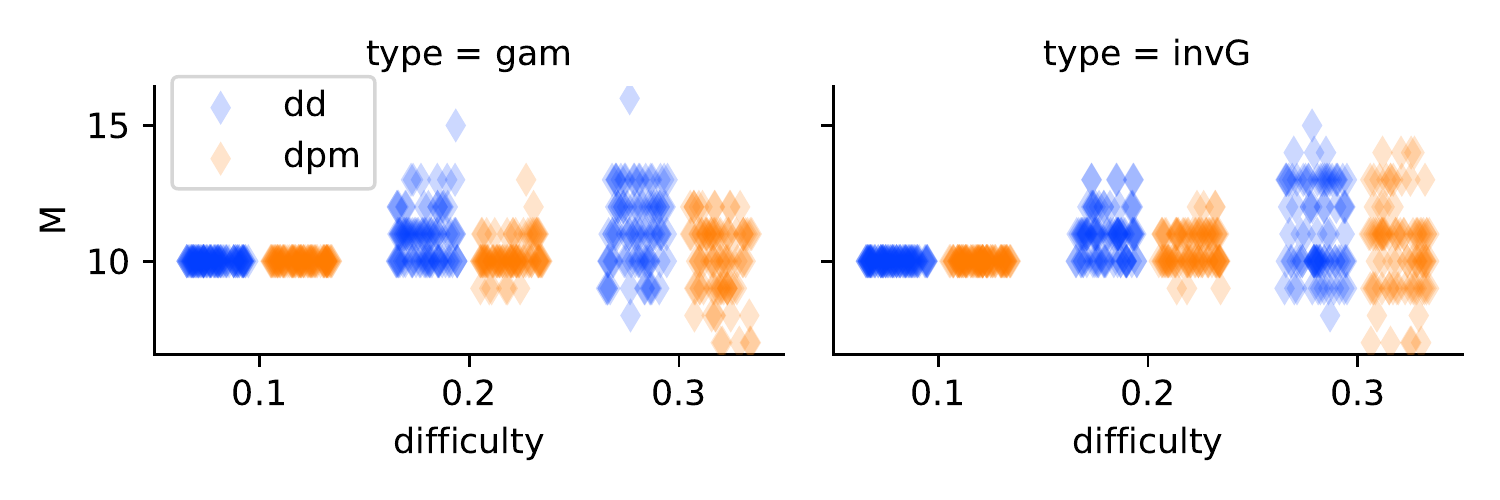}
    \caption{
    Comparison between finite (Dirichlet distribution) and infinite (Dirichlet process mixture) model 
    in estimation number of clusters for unbaranced population data.
    }
    \label{fig:pop}
\end{figure}

The relationship between ELBO and ARI
for the proposed method shows a strong correlation (Fig.~\ref{fig:lb}).
This shows that selecting a large ELBO result
from multiple output with different initial conditions
yields a better performance.
In the relationship between the ELBO and the ARI,
There was no difference between the finite (Dirichlet distribution) model
and the infinite (Dirichlet process mixture) model (Fig.~\ref{fig:lb}).
In the non-uniform population case,
the estimate of the number of clusters
is slightly worse than that in the uniform case (Fig.~\ref{fig:num}).
The estimation by the infinite model is a slightly better estimate
than that of the finite model,
but it is not significantly different (Fig.~\ref{fig:num}).

\section{Discussion}
We proposed a variational Bayesian clustering method
for heavy tailed and/or asymmetric data
based on a variational Bayes algorithm
for NIG mixture models as an improvement of an existing model.
In addition to the finite mixture model with Dirichlet distributions,
Dirichlet process mixture were also derived.
In the evaluation by artificial data,
the proposed method performed much better than
the Gaussian and existing NIG distribution models,
especially in the case of normality $\lambda$ small.

In this paper, in addition to the infinite and finite implementations
we have two prior distributions of non-normality:
the gamma distribution and the inverse gamma distribution.
None of the implementation combinations showed significant differences for artificial data.
If we have some prior knowledge of the data,
we can set each hyperparameter more appropriately.
The adjustment of hyperparameters by empirical Bayesian methods
is also a topic worthy of further study.

\section*{Acknowledgements}
This work was supported by JSPS KAKENHI Grant Number 19K12104.

\appendix

\section{Distributions}

In this section,
we use
the modified Bessel function of the third kind 
$K_c\func{\cdot}$
of order $c$ 
the gamma function $\Gamma$,
the $D$-dimensional gamma function $\Gamma_D$
and 
the digamma function $\psi$.

\subsection{Definitions}

The generalized inverse Gaussian distribution is defined as
\begin{gather}
\log \mathcal{N}_*^{-1}\prob{x}{a, b, c}
= \Delta\func{a, b, c}
+ \left(c - 1\right) \log x - \frac{a}{2}  - \frac{b}{2} x^{-1},
\\
\Delta\func{a, b, c}
=  - \log 2 + \frac{c}{2} \log \frac{a}{b} - \log K_c\func{\sqrt{ab}}.
\end{gather}
The generalized inverse Gaussian with $c=-\frac{1}{2}$
is inverse Gaussian distribution
\begin{equation}
\begin{aligned}
    \log \mathcal{N}_*{-1}\prob{x}{a, b, -\frac{1}{2}}
    &=
    \log \mathcal{N}^{-1}\prob{x}{\mu=\sqrt{\frac{b}{a}}, \lambda=\sqrt{ac}}
    \\
    &=
    - \frac{1}{2} \log 2 \pi
    + \frac{1}{2} \log \lambda
    + \lambda
    - \log \mu
    - \frac{3}{2} \log \frac{x}{\mu}
    - \frac{\lambda}{2} \frac{x}{\mu} - \frac{\lambda}{2} \frac{\mu}{x}
    ,
\end{aligned}
\end{equation}
and the generalized inverse Gaussian with $b=0$
is Gamma distribution
\begin{equation}
\begin{aligned}
    \log \mathcal{N}_*{-1}\prob{x}{a, 0, c}
    &=
    \log \mathcal{G}\prob{x}{\alpha=\sqrt{c}, \beta=\frac{a}{2}}
    \\
    &=
    - \log \Gamma\func{\alpha}
    + \alpha \log \beta
    - \func{\alpha-1} \log x
    - \beta x
    .
\end{aligned}
\end{equation}
Dirichlet, Beta, Wishart and normal distribution are respectively difined as
\begin{gather}
    \log \mathcal{D}\prob{\alpha_1, \cdots \alpha_M}{l_1, \cdots, l_M}
    = \log \Gamma\func{\sum_{j=1}^Ml_j} - \sum_{j=1}^M\log\Gamma\func{l_j}
    + \sum_{j=1}^M \func{l_j - 1}\log \alpha_j,
    \\
    \log \mathcal{B}\prob{x}{\alpha, \beta}
    = \Gamma\func{\alpha + \beta} - \Gamma\func{\alpha} - \Gamma\func{\beta}
    + \func{\alpha-1}\log x + \func{\beta-1}\log \func{1-x},
    \\
    \log \mathcal{W}\prob{x}{\alpha, \beta}
    =
    \frac{\alpha}{2} \log \det \frac{\beta}{2}
    - \log \Gamma_D\func{\frac{\alpha}{2}}
    + \frac{\alpha - D - 1}{2} \log \det x  - \frac{1}{2} \tr \beta x,
    \\
    \log \mathcal{N}\prob{x}{\mu, \tau}
    = -\frac{d}{2} \log 2\pi + \frac{1}{2} \log \det \tau
    - \frac{1}{2}\tr \tau \func{x - \mu}\func{x - \mu}^\top.
\end{gather}

\subsection{The mixing density}

In the definition of the proposed model,
the mixing density obey the inverse Gaussian which mean is 1:
\begin{equation}
\begin{split}
\log \mathcal{N}^{-1}\prob{y}{y_0=1, \lambda}
&= \log \mathcal{N}_*^{-1}\prob{y}{\lambda, \lambda, -\frac{1}{2}}
\\
&= -\frac{1}{2}\log 2\pi
+ \frac{1}{2}\log\lambda
+ \lambda
- \frac{3}{2} \log y
- \frac{\lambda}{2} y
- \frac{\lambda}{2} y^{-1}.
\end{split}
\end{equation}
And the expectation values of the posterior
$q\prob{y}{z} = \mathcal{N}_*^{-1}\prob{y}{a, b, c}$
are calculated as
\begin{align}
    \expect{y}{\mathcal{N}_*^{-1}\prob{y}{a, b, c}}
    &= \sqrt{\frac{b}{a}}\frac{K_{c+1}\func{\sqrt{ab}}}{K_c\func{\sqrt{ab}}},
    &
    \expect{y^{-1}}{\mathcal{N}_*^{-1}\prob{y}{a, b, c}}
    &= \sqrt{\frac{a}{b}}\frac{K_{c-1}\func{\sqrt{ab}}}{K_c\func{\sqrt{ab}}}.
\end{align}

\subsection{Expectations for posteriors}
For Dirichlet distribution models, the conjugate prior of $\alpha$ is
$q\func{\alpha} = \mathcal{D}\prob{\alpha}{l}$
and the expectation values are
\begin{gather}
    \expect{\log \alpha_j}{\mathcal{D}\prob{\alpha}{l}}
    = \psi\func{l_j} - \psi\func{\textstyle \sum_{j'=1}^M l_{j'}}.
\end{gather}
For DPM, the conjugate prior of $\gamma$ is beta distribution
$q\func{\gamma} = \mathcal{B}\prob{\gamma}{l ,r}$
and the expectation values are
\begin{gather}
    \expect{\log \gamma_j}{\mathcal{B}\prob{\gamma}{l, r}}
    = \psi\func{l} - \psi\func{l + r},
    \\
    \expect{\log \func{1-\gamma_j}}{\mathcal{B}\prob{\gamma}{l, r}}
    = \psi\func{r} - \psi\func{l + r}
    .
\end{gather}

The posterior of $\lambda$ obey the generalized inverse Gaussian
$q\func{\lambda} = \mathcal{N}_*^{-1}\prob{\lambda}{f,g,h}$
and the expectation values are
\begin{gather}
    \expect{\lambda}{\mathcal{N}_*^{-1}\prob{\lambda}{f, g, h}}
    = \sqrt{\frac{g}{f}}\frac{K_{h+1}\func{\sqrt{fg}}}{K_h\func{\sqrt{fg}}},
    ,\\
    \expect{\log \lambda}{\mathcal{N}_*^{-1}\prob{\lambda}{f,g,h}}
    = \log \sqrt{\frac{g}{f}}
    + \frac{\partial \log K_h}{\partial h}\func{\sqrt{fg}}
\end{gather}
In the case that $\lambda$ obey Gamma distribution $q\func{\lambda} = \mathcal{N}_*^{-1}\prob{\lambda}{f,0,h} = \mathcal{G}\prob{\lambda}{h, f/2}$,
which is the special case of the generalized invverse Gaussin with $g=0$, 
the expectation values are
\begin{gather}
    \expect{\lambda}{\mathcal{G}\prob{\lambda}{h, f/2}}
    = \frac{2h}{f}
    ,\\
    \expect{\log \lambda}{\mathcal{G}\prob{\lambda}{h, f/2}}
    = \psi\func{h} - \log f + \log 2
\end{gather}

The posterior of $\tau$ obey Wishart distribution
$q\func{\tau} = \mathcal{W}\prob{\tau}{s, t}$
and the expectation values are
\begin{gather}
    \expect{\tau}{\mathcal{W}\prob{\tau}{s, t}}
    = s t^{-1}
    ,\\
    \expect{\log\det\tau}{\mathcal{W}\prob{\tau}{s, t}}
    = \psi_d\func{\frac{s}{2}} - \log\det\frac{t}{2}
    .
\end{gather}
The posterior of $\mu$ and $\beta$ obey Normal distribution:
\begin{equation}
    q_\mu\prob{\mu, \beta}{\tau} =
    \mathcal{N}\prob{
    \begin{matrix}
    \mu \\ \beta
    \end{matrix}
    }{
    \begin{matrix}
    m \\ n
    \end{matrix}
    ,\;
    \begin{matrix}
    u\tau & w\tau \\ w\tau & v\tau
    \end{matrix}
    }.
\end{equation}
and the expectation values are
\begin{gather}
    \expect{\mu}{q\prob{\mu,\beta}{\tau}}
    = m
    ,\quad
    \expect{\beta}{q\prob{\mu,\beta}{\tau}}
    = n
    ,\\
    \expect{\tau \mu \mu^\top}{q\prob{\mu,\beta}{\tau}}
    = u^{-1} I
    ,\quad
    \expect{\tau \mu \beta^\top}{q\prob{\mu,\beta}{\tau}}
    = w^{-1} I
    ,\quad
    \expect{\tau \beta \beta^\top}{q\prob{\mu,\beta}{\tau}}
    = v^{-1} I
    .
\end{gather}

\subsection{Prior setting}

The inverse Gaussian distribution is a special case of generalized inverse Gaussian with $c=-1/2$ and described by mean $\mu$ and shape $\lambda$ parameter
\begin{gather}
%
\mathcal{N}_*^{-1}\prob{\lambda}{f, g, -\frac{1}{2}}
=
\mathcal{N}^{-1}\prob{\lambda}{\sqrt{\frac{g}{f}}, \sqrt{f g}},
\\
\lambda_0 = 
\expect{\lambda}
{\mathcal{N}_*^{-1}\prob{\lambda}{f, g, -\frac{1}{2}}}
= \sqrt{\frac{g}{f}},
\quad
\nu_\lambda
= \lambda_0^2
\expectsup{\func{\lambda-\lambda_0}^2}
{\mathcal{N}_*^{-1}\prob{\lambda}{f, g, -\frac{1}{2}}}{-1}
= \sqrt{fg}
\end{gather}

The gamma distribution is also special case of generalized inverse Gaussian with $c=0$
and described by shape $\alpha$ and rate $\beta$ parameter
\begin{gather}
\mathcal{N}_*^{-1}\prob{x}{f, 0, h}
=
\mathcal{G}\prob{x}{h, \frac{f}{2}},
\\
\lambda_0 = 
\expect{\lambda}
{\mathcal{N}_*^{-1}\prob{\lambda}{f, 0, h}}
= \frac{2h}{f},
\quad
\nu_\lambda
= \lambda_0^2
\expectsup{\func{\lambda-\lambda_0}^2}
{\mathcal{N}_*^{-1}\prob{\lambda}{f, 0, h}}{-1}
= h
\end{gather}

\section{Difference between the previous and the proposed model}

Main difference between the previous model \citep{Subedi2014} and the proposed model
is limitation to the mixing density.
The inverse Gaussian distribution has the mean and the shape parameters.
The previous model fix the shape parameter
and it conclude the truncated normal distribution $\mathcal{N}_{>0}$
for the conjugate prior.

\begin{figure}
    \centering
    \includegraphics{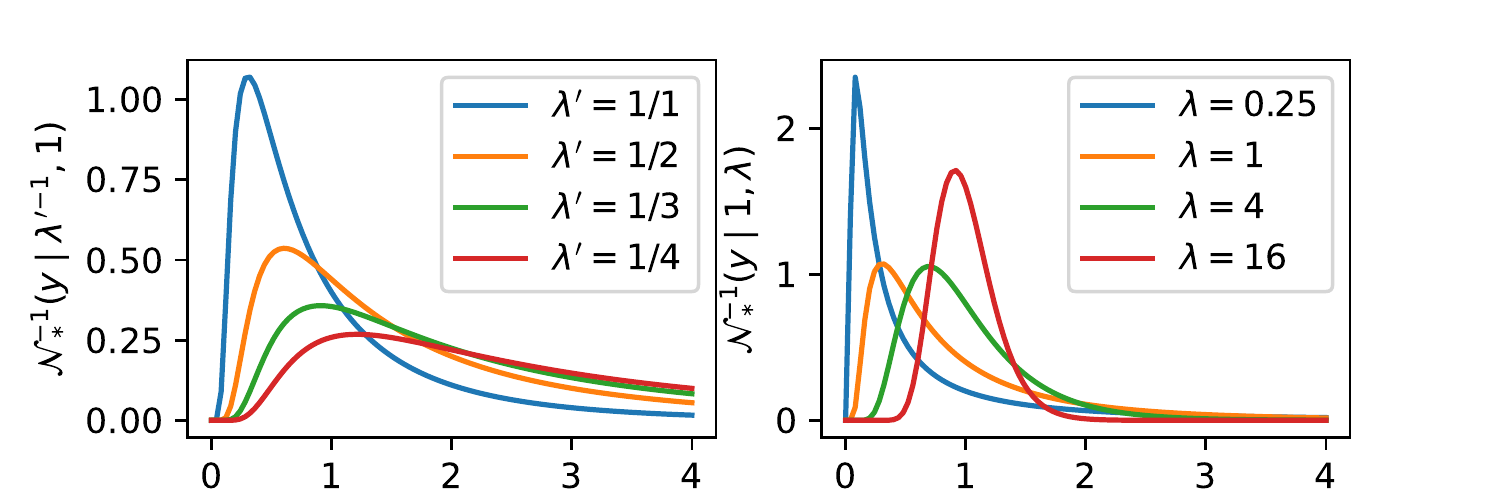}
    \caption{Probability density function of the mixing density of the previous model (left) and the proposed model (right).}
    \label{fig:invgauss}
\end{figure}

They defined that the probability of the mixing density $y$ obey
\begin{equation}
    p\func{y} = \mathcal{N}^{-1}\prob{y}{\frac{1}{\lambda}, 1}
    .
\end{equation}
In this condition, the terms of $\lambda'$ in the expectation of joint model
can be written as
\begin{equation}
\expect{\log p\prob{x, y, z}{\alpha,\lambda,\mu,\beta,\tau}}{q_\zeta}
=
\sum_{j=1}^M \left\{
- \frac{Z^+_j}{2} {\lambda_j}^2
+ Z^*_j \lambda_j
\right\} + \cdots
\end{equation}
And the condugate prior and the update rules in M-step can described as
\begin{equation}
    p\func{\lambda_0} = \mathcal{N}_{>0}\prob{\lambda_j}{f_0, g_0}
    ,\quad
    q\func{\lambda_j} = \mathcal{N}_{>0}\prob{\lambda_j}{f_j, g_j}
\end{equation}
and
\begin{equation}
    f_j = \frac{f_0g_0 + Z_j^*}{g_0 + Z_j^+}
    ,\quad
    g_j = g_0 + Z_j^+
    .
\end{equation}

In E-step, Eq.~\eqref{ab} and \eqref{rho} are replaced by the following equations:
\begin{equation}
\begin{aligned}
a_j &=
\expect{\lambda_j^2}{q_\theta}
+ \tr \expect{\tau_j\beta_j\beta_j^\top}{q_\theta}
+ \tr \expect{\tau_j}{q_\theta}\expect{\beta_j}{q_\theta}
\expectsup{\beta_j}{q_\theta}{\top}
,\\
b_{ij} &=
1 + \tr \expect{\tau_j\mu_j\mu_j^\top}{q_\theta}
+ \tr \expect{\tau_j}{q_\theta}
\func{x_i-\expect{\mu_j}{q_\theta}}
\func{x_i-\expect{\mu_j}{q_\theta}}^\top
\end{aligned}
\end{equation}
and
\begin{multline}
    \log \rho_{ij} =
- \frac{D+1}{2}\log 2\pi
+ \expect{\log\alpha_j}{q_\theta}
+ \expect{\lambda_j}{q_\theta}
+ \frac{1}{2} \expect{\log\det\tau_j}{q_\theta}
\\
- \tr \expect{\tau_j\mu_j\beta_j}{q_\theta}
+ \tr \expect{\tau_j}{q_\theta}\func{x_i-\expect{\mu_j}{q_\theta}}\expectsup{\beta_j}{q_\theta}{\top}
- \log \Delta\func{a_j, b_{ij}, c}.
\end{multline}\label{rho2}

\bibliography{nig}

\end{document}